\documentclass[letterpaper, 10 pt, conference]{ieeeconf}  %

\IEEEoverridecommandlockouts                              %

\usepackage[noend]{algpseudocode}
\usepackage{algorithm}
\usepackage{amsmath}
\usepackage{amssymb}
\usepackage[normalem]{ulem}
\usepackage{wrapfig}
\usepackage{bm}
\usepackage{xcolor}
\usepackage{graphicx}
\usepackage{hyperref}
\usepackage{booktabs}
\usepackage{caption}
\usepackage[frozencache,cachedir=.]{minted}
\setminted{escapeinside=||}

\newcommand{\var}[1]{\textit{#1}}
\newcommand{\proc}[1]{\textsc{#1}}

\newcommand{\kw}[1]{{\bf #1}}

\newcommand{\True}[0]{\kw{True}}

\newcommand{\None}[0]{\kw{None}}

\newcommand{\R}{\mathbf{R}}

\newcommand{\SE}[1]{\mathrm{SE}(#1)}

\usepackage{amsthm}
\theoremstyle{plain}
\theoremstyle{definition}\newtheorem{defn}{Definition}
\theoremstyle{plain}
\theoremstyle{plain}

\usepackage{listings}
\definecolor{deepblue}{rgb}{0,0,0.5}
\definecolor{deepred}{rgb}{0.6,0,0}
\definecolor{magenta}{rgb}{1.0,0,1.0}
\definecolor{deepgreen}{rgb}{0,0.5,0}
\definecolor{textblue}{rgb}{.2,.2,.7}
\definecolor{textred}{rgb}{0.54,0,0}
\definecolor{textgreen}{rgb}{0,0.43,0}
\definecolor{es-blue}{rgb}{0.1372,0.666,1}
\definecolor{lightgraybg}{RGB}{247,247,247}  %

\lstdefinestyle{pythonstyle}{
    language=Python, 
    breaklines=true,
    basicstyle=\ttfamily\small,
    emphstyle=\bfseries\color{deepred}, 
    emph={forward,forward_v},         
    numbers=left,
    numberstyle=\tiny\color{gray},
    stepnumber=1,
    numbersep=12pt,
    tabsize=2,
    stringstyle=\color{textgreen},
    frame=none,                    
    columns=fullflexible,
    keepspaces=true,
    xleftmargin=\parindent,
    showstringspaces=false,
    commentstyle=\color{deepgreen},
    keywordstyle=\color{es-blue},
    backgroundcolor=\color{lightgraybg}  %
}

\newcommand{\python}[1]{\mintinline{python}{#1}}
\newcommand{\kwarg}[1]{\textcolor{deepblue}{\textit{#1}}}
\newcommand{\class}[1]{\textcolor{deepblue}{\textbf{#1}}}
\newcommand{\approach}{ScheduleStream} %

\newcommand{\caelan}[1]{}
\newcommand{\fabio}[1]{\textcolor{green}{}}
\newcommand{\todo}[1]{}

\title{\LARGE \bf
ScheduleStream: Temporal Planning with Samplers for \\
GPU-Accelerated Multi-Arm Task and Motion Planning \& Scheduling
}

\author{Caelan Garrett$^{1}$ and Fabio Ramos$^{1,2}$%
\thanks{$^{1}$NVIDIA Research Seattle Robotics Lab (SRL)  } %
\thanks{$^{2}$University of Sydney, {\tt\footnotesize \{cgarrett,ftozetoramos\}@nvidia.com}}%
}

\begin{document}

\maketitle
\thispagestyle{empty}
\pagestyle{empty}

\begin{abstract}
Bimanual and humanoid robots are appealing
because of their human-like ability to leverage multiple arms to efficiently complete tasks.
However, controlling multiple arms at once is computationally challenging due to the growth in the hybrid discrete-continuous action space.
Task and Motion Planning (TAMP) algorithms can efficiently plan in hybrid spaces but generally produce plans, where only one arm is moving at a time, rather than schedules that allow for parallel arm motion.
In order to extend TAMP to produce schedules,
we present \approach{}, the first general-purpose framework for planning \& {\em scheduling} with sampling operations.
\approach{} models temporal dynamics using hybrid durative actions, 
which can be started asynchronously and persist for a duration that's a function of their parameters. %
We propose domain-independent algorithms that solve \approach{} problems without any application-specific mechanisms.
We apply \approach{} to Task and Motion Planning \& Scheduling (TAMPAS), where we use GPU acceleration within samplers to expedite planning.
We compare \approach{} algorithms to several ablations 
in simulation and find that they produce more efficient solutions. %
We demonstrate \approach{} on several real-world bimanual robot tasks at \url{https://schedulestream.github.io}.
\end{abstract}
\vspace{-0.5cm}

\section{INTRODUCTION}

There's been a spike in demand for bimanual and humanoid robots to be used to automate tasks in both industrial and home settings, in place of traditional manipulators.
However, programming these robots to take full advantage of their multiple arms through simultaneous manipulation
remains challenging.
Imitation learning from human teleoperation demonstrations can be an effective control approach 
but requires an enormous amount of data, and thus human effort, even for narrowly-defined tasks~\cite{robomimic2021}. %
In contrast, automated robot planning, such as Task and Motion Planning (TAMP)~\cite{Garrett2021}, is able to plan behaviors that generalize across a broad distribution of tasks through composing a set of engineered or learned primitive actions.
Additionally, planning can be used to develop imitation learning systems by serving as an automated training data generator~\cite{dalal2023imitating}.

\begin{figure}
    \centering
    \includegraphics[width=\columnwidth]{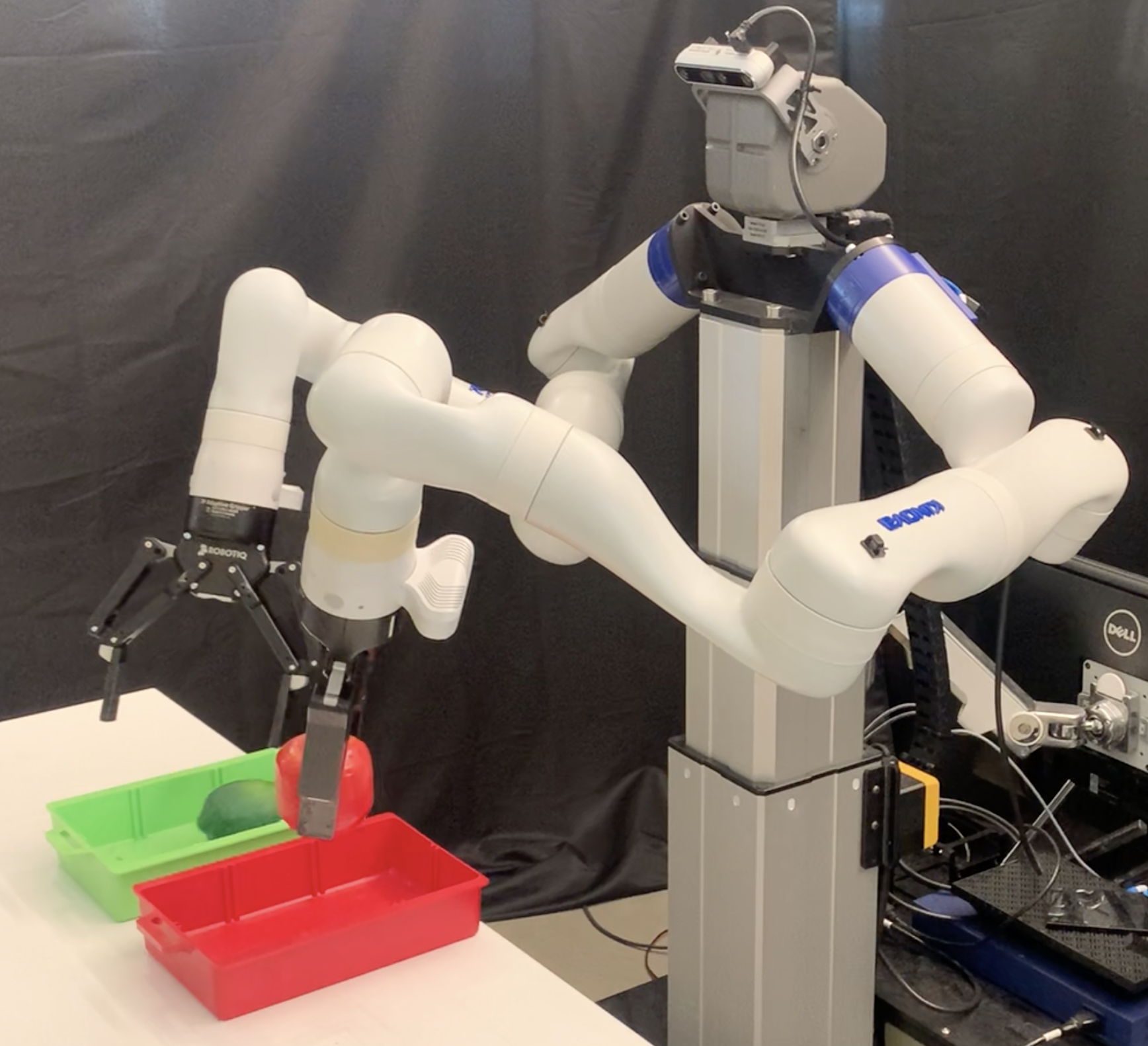}
    \caption{\small{
    \textbf{Real-World Demonstration.} Using \approach{}, a bimanual robot solves for a schedule to sort the apple into the red bin and the lime in the green bin. \approach{} algorithms automatically select which arm to use for which object based on kinematics and execute the actions asynchronously in parallel. %
    }}
    \vspace{-0.75cm}
    \label{fig:teaser}
\end{figure}

In order to safely and productively control robots, TAMP algorithms must make discrete decisions, such as which arm to use to manipulate which object and how, as well as continuous decisions in the form of the motion and other parameters that lead to executable robot actions.
Consider the bimanual robot in Figure~\ref{fig:teaser}, which is tasked with sorting the two fruits into the two bins. 
In order to maximize execution efficiency, a TAMP algorithm should plan to use both arms in parallel, but critically, it must identify which arms can reach which objects by reasoning about reachability.

Most existing TAMP approaches~\cite{Garrett2021} are only able to produce {\em plans}, serial sequences of actions, as opposed to {\em schedules}, sets of timed actions that can be executed in serial or parallel.
As a result, they will only produce solutions that have one arm moving at a time, preventing them from using both arms in parallel to minimize schedule {\em makespan}, or duration.
The existing work that incorporates multi-arm parallel manipulation typically is application-specific~\cite{chen2022cooperative,hartmann2022long,huang2025apex}, addresses a strict subclass of TAMP~\cite{shome2019anytime,wei2025hierarchical}, or has limitations in its ability to leverage parallelism~\cite{pan2021general,zhang2023multi,lee2025lazy}.

In this work, we seek to address full Task and Motion Planning \& Scheduling (TAMPAS), where solutions are schedules
of timed hybrid actions. 
Unlike all prior work, we accomplish this by first introducing \approach{}, a new domain-independent framework for planning \& scheduling with samplers.
We implement this language in the programming language Python, %
allowing external procedures such as GPU-accelerated collision checkers and inverse kinematics solvers
to flexibly be invoked by \approach{} algorithms.
We propose domain-independent \approach{} algorithms that alternate between scheduling and sampling phases, which solve each scheduling subproblem through planning the start and end of each action.
In simulated experiments, we show that these algorithms have higher success rates than strictly hierarchical algorithms that schedule {\em then} motion plan~\cite{chen2022cooperative} and lower makespans than serial-only planning algorithms.
\textbf{The contributions of this paper are:}
\begin{itemize}
    \item \approach{}: the first domain-independent {\em temporal} planning language with support for procedural functions and samplers ({\it i.e.} streams) to enable planning \& scheduling for mixed discrete-continuous systems. %
    \item Novel \approach{} algorithms that {\em lazily} solve problems while minimizing schedule makespan.
    \item An application of \approach{} to Task and Motion Planning \& Scheduling (TAMPAS), which leverages GPU acceleration for {\em parallelized} sampling.
    \item Simulated and real-world experiments and demonstrations across multiple multi-arm platforms, including a real-world bimanual robot.
\end{itemize}

\section{RELATED WORK}\label{sec:related}

We build on prior work in single-arm Task and Motion Planning (TAMP) as well as multi-arm manipulation. %

\subsection{Task and Motion Planning} %

There are a number of existing approaches for traditional TAMP that can efficiently produce single-arm plans but are unable to produce multi-arm schedules~\cite{Garrett2021}.
Several approaches apply TAMP to humanoid robots~\cite{HauserIJRR11,grey2016humanoid}; however, they only address single-arm loco-manipulation rather than bimanual manipulation. %
cuTAMP~\cite{shen2025cutamp} proposed leveraging GPU parallelization for single-arm sampling and gradient-based TAMP.
The closest to our approach is PDDLStream~\cite{garrett2020PDDLStream}, which also is a domain-independent planning framework for hybrid systems. %
Key differences from \approach{} are that PDDLStream describes domains using text ({\it i.e.} PDDL) instead of code ({\it i.e.} Python) and does not support temporal, functional, or GPU-batched planning.

\subsection{Multi-Arm Manipulation} %

Several prior works plan for multiple arms in automated assembly, an application of TAMP that makes various assumptions about task allocations, monotonicity, single object goals
~\cite{chen2022cooperative,hartmann2022long,huang2025apex}.
Many of these approaches also factor discrete and continuous planning into a strict hierarchy and will fail to solve problems that are not downward refinable~\cite{bacchus1991downward}, problems where some high-level plans aren't viable.

Several approaches address {\em synchronous} multi-arm manipulation~\cite{pan2021general,zhang2022mip,zhang2023multi}, where each robot takes an action that starts and ends simultaneously according to a global clock.
Such synchronicity requirements restrict the space of schedules, limiting opportunities to minimize schedule makespan.
Still, many traditional TAMP systems can implement this strategy by treating the arms collectively as a single composite robot~\cite{Toussaint2017Multi-boundDomains,driess2021learning}.
However, these strategies discard factoring and often sparse interaction among robots, leading to combinatorial growth in their search spaces.

Two approaches investigate multi-arm handover tasks.
Shome {\it et al.}~\cite{shome2019anytime} extend dRRT*~\cite{shome2020drrt}, a sampling-based multi-robot motion planner, and 
Wei {\it et al}~\cite{wei2025hierarchical} combine Linear Temporal Logic (LTL) and Graph of Convex Sets (GCS) in a hierarchical manner.
However, these planners would need to be continuously extended to address more general manipulation tasks, for example, ones where objects aren't limited to a pre-determined set of placements.

Hartmann and Toussaint~\cite{hartmann2023towards} develop a rescheduling algorithm that 
greedily reschedules an inputted serial plan subject to conflicts to locally minimize its makespan.
The Decomposable State Space Hypergraph (DaSH) line-of-work~\cite{motes2023hypergraph,lee2025lazy} also presents several algorithms 
that explicitly factor problems into separate planning spaces per robot and later resolve conflicts to produce a schedule. %
Unlike \approach{}, these algorithms are application-specific and don't directly minimize schedule makespan.

\section{SCHEDULESTREAM LANGUAGE}\label{sec:language} %

\begin{figure*}[ht]
    \centering
    \includegraphics[width=0.32\textwidth]{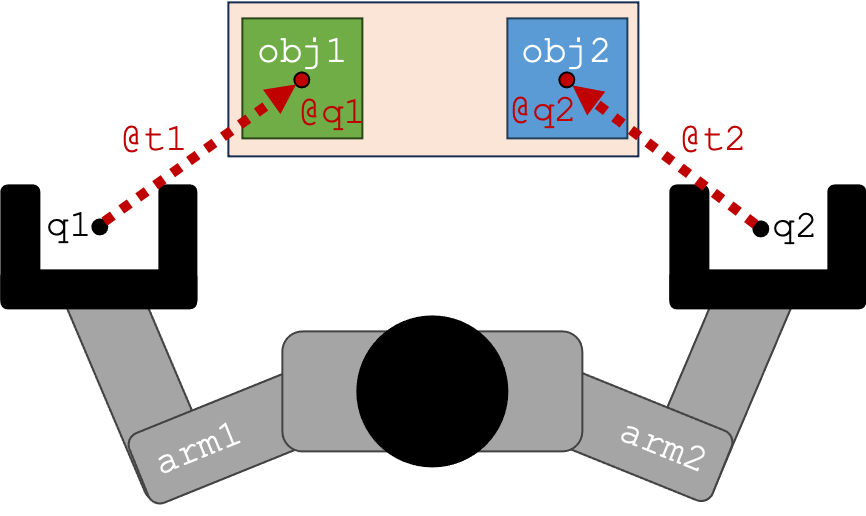}
    \includegraphics[width=0.32\textwidth]{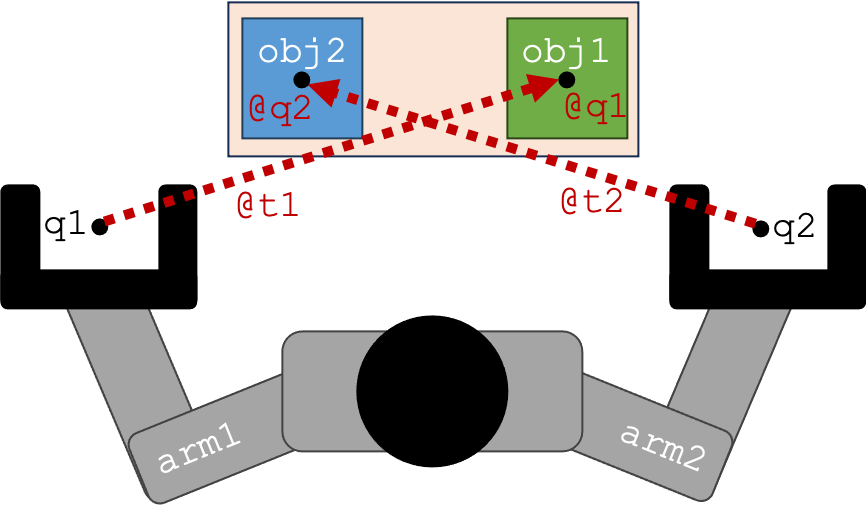}
    \includegraphics[width=0.32\textwidth]{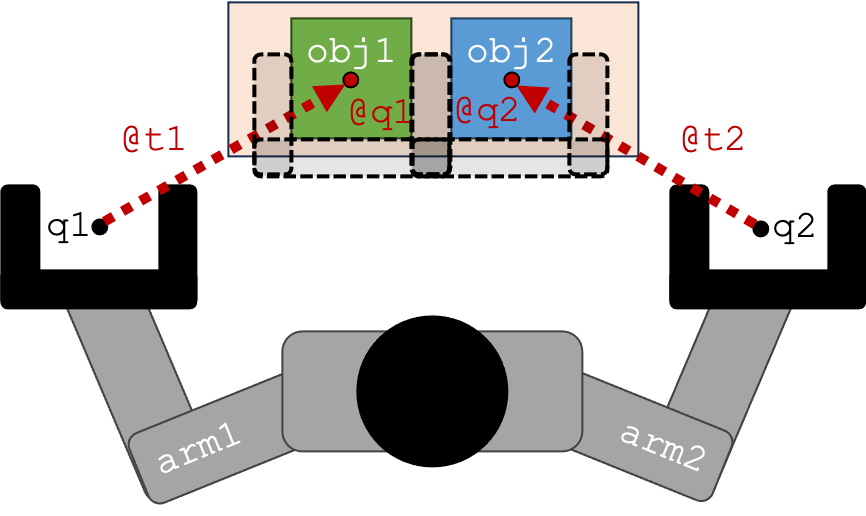}
    \caption{\small{\textbf{Illustrative Examples}. Three example bimanual TAMPAS problems where the goal is for arm \python{arm1} to hold object \python{obj1} and arm \python{arm2} to hold object \python{obj2}. {\it Problem 1} ({\it left}): the arms can pick their objects in parallel. {\it Problem 2} ({\it center}): the arms can only pick their objects sequentially. {\it Problem 3} ({\it right}): one arm must retreat after picking its object in order to make space for the other.
    }}
    \label{fig:bimanual}
    \vspace{-2em}
\end{figure*}

We start by introducing \approach{}, a new general-purpose language for planning \& scheduling with continuous sampling operations.
We will use a simplified bimanual manipulation domain as a running example.
Figure~\ref{fig:bimanual} visualizes three cartoon problems in this domain, where there's a bimanual robot with two arms, \python{arm1} and \python{arm2}, and the goal in each problem is for \python{arm1} to hold object \python{obj1} and \python{arm2} to hold object \python{obj2}.
We build \approach{} entirely in the Python programming language, which, unlike prior works~\cite{mcdermott1998pddl,garrett2020PDDLStream}, enables direct interfacing between the declarative and procedural language aspects.

We consider a universe of possibly infinitely many {\em constants}, atomic values. %
In our bimanual domain, there are 1) {\em discrete} constants such as first-class objects for the arms \python{arm1}, \python{arm2} and the objects \python{obj1}, \python{obj2} and 2) infinitely-many {\em continuous} constants in the form of (PyTorch) tensors for arm configurations \python{q} $\in \R^d$, object placement poses \python{p} $\in \SE{3}$, and object grasp poses \python{g} $\in \SE{3}$.

Similar to Functional STRIPS~\cite{geffner2000functional}, we describe states and actions using declarative {\em functions}, relations between a tuple of input constants and a single output constant.
This subsumes predicate-based languages, such as STRIPS~\cite{Fikes71} and Planning Domain Definition Language (PDDL)~\cite{mcdermott1998pddl}, which only consider Boolean functions.
Each function \python{|\class{Function}|}, or predicate \python{|\class{Predicate}|}, is defined by a string of input {\em parameters} (e.g. \python{"?x1 ... ?xN"}) and optionally a domain {\em condition} \python{|\kwarg{cond}|} that specifies predicates that the input parameters are guaranteed to satisfy.
For example, let \python{At} be a function from an arm \python{"?arm"} parameter that satisfies  typing predicate \python{Arm("?arm")} to its current configuration.
And let \python{Conf("?arm ?q")} be \python{True} if \python{"?q"} is a valid configuration for arm \python{"?arm"} in some state.

\begin{footnotesize}
\begin{minted}{python}
At = |\class{Function}|("?arm", |\kwarg{cond}|=[Arm("?arm")])
Arm = |\class{Predicate}|("?arm"); Conf = |\class{Predicate}|("?arm ?q")
\end{minted}
\end{footnotesize}
States are assignments to the functions in the domain.
For example, the {\em assignment} \python{At(arm1)<=q1} declares that function \python{At(arm1)} with argument \python{arm1} currently evaluates to configuration \python{q1}.
We make a version of the {\em closed-world} assumption in that any undeclared function and predicate evaluate to \python{None} and \python{False} respectively.
We can also {\em test} whether a function evaluates to a particular constant using an equation \python{At(arm1)==q1}, which evaluates to either \python{True} or \python{False}.
In our Python implementation, we override the \python{<=} and \python{==} operations for modeling syntactic ease.

Often, some functions are {\em static}, meaning that their values are fixed over time, and others are {\em fluent}, meaning that their values change over time as actions are applied.
In our bimanual domain, the static object-related predicates are \python{Object("?obj")} and \python{Placement("?obj ?p")}, which hold if \python{"?obj"} is an object and \python{"?p"} is a valid placement pose for object \python{"?obj"} respectively.

\begin{footnotesize}
\begin{minted}{python}
Object = |\class{Predicate}|("?obj")
Placement = |\class{Predicate}|("?obj ?p")
\end{minted}
\end{footnotesize}
The fluent object-related functions are \python{Holding("?arm")} which indicates which object the \python{"?arm"} is holding, \python{Attached("?obj")} which indicates which arm the \python{"?obj"} is attached to, and \python{Pose("?obj")} the current pose of \python{"?obj"}. %

\begin{footnotesize}
\begin{minted}{python}
Holding = |\class{Function}|("?arm", |\kwarg{cond}|=[Arm("?arm")])
Attached = |\class{Function}|("?obj", |\kwarg{cond}|=[Object("?obj")])
Pose = |\class{Function}|("?obj", |\kwarg{cond}|=[Object("?obj")])
\end{minted}
\end{footnotesize}

Returning to the problems in Figure~\ref{fig:bimanual}, let \python{q1}, \python{q2} $\in R^{d}$ be the start configurations for arms \python{arm1}, \python{arm2} and \python{p1}, \python{p2} $\in \SE{3}$ be the start placements for objects \python{obj1}, \python{obj2}.
Then, the initial state function assignments are:

\begin{footnotesize}
\begin{minted}{python}
initial_state = [At(arm1)<=q1, At(arm2)<=q2, 
  Holding(arm1)<=None, Holding(arm2)<=None,
  Pose(obj1)<=p1, Pose(obj2)<=p2, 
  Attached(obj1)<=None, Attached(obj2)<=None, ...]
\end{minted}
\end{footnotesize}
and the goal is the logical conjunction expressed as a list that \python{arm1} is holding \python{obj1} and \python{arm2} is holding \python{obj2}.

\begin{footnotesize}
\begin{minted}{python}
goal = [Holding(arm1)==obj1, Holding(arm2)==obj2]
\end{minted}
\end{footnotesize}

\subsection{Instantaneous Actions} \label{sec:instantaneous}

In order to describe {\em actions}, operations that change the current state, we adopt a precondition/effect representation similar to that of STRIPS~\cite{Fikes71} or PDDL~\cite{mcdermott1998pddl}.
An action \python{|\class{Action}|} has a string of input {\em parameters} \python{|\kwarg{params}|}, a list of equality {\em conditions} \python{|\kwarg{cond}|}, and a list of assignment {\em effects} \python{|\kwarg{eff}|}.
In order to apply an action, the conditions must hold in the current state.
After applying an action, the state is updated according to the effect assignments.

In our running example, we model \python{pick} and \python{place} as actions that cause an instantaneous change in the parent of object \python{"?obj"} in the scene graph.
The \python{pick} action has parameters for the arm \python{"?arm"}, pick configuration \python{"?q"}, object \python{"?obj"}, before placement pose \python{"?p"}, and after grasp pose \python{"?g"}.
Its conditions are that the parameters comprise a valid kinematic solution \python{Kin} (see Section~\ref{sec:streams}), \python{"?obj"} isn't initially attached, \python{"?obj"} is initially at placement pose \python{"?p"}, \python{"?arm"} isn't initially holding anything, and \python{"?arm"} is currently at configuration \python{"?q"}.

\begin{footnotesize}
\begin{minted}{python}
pick = |\class{Action}|(|\kwarg{params}|="?arm ?q ?obj ?g ?p",
  |\kwarg{cond}|=[Kin("?arm ?q ?obj ?g ?p"),
        Attached("?obj")==None, Pose("?obj")=="?p", 
        Holding("?arm")==None, At("?arm")=="?q"],
  |\kwarg{eff}|=[Holding("?arm")<="?obj", Pose("?obj")<="?g", 
       Attached("?obj")<="?arm"])
\end{minted}
\end{footnotesize}
\python{place} is defined similarly, as the reverse of \python{pick}.

\subsection{Durative Actions} \label{sec:durative}

Existing TAMP frameworks generally only support sequential plans~\cite{garrett2020PDDLStream}, where solutions are sequences of actions.
We wish to beyond this to support {\em schedules} as solutions, where actions can asynchronously overlap in time.
To do this, we generalize the concept of an instantaneous action to a durative {\em action}~\cite{Fox03pddl2.1:an} that can be {\em started} at a point in time and {\em ended} after a certain duration.
Each start and end {\em event} causes a change to the state.
In order to correctly execute durative actions, certain conditions must hold before the start event, while the action is ongoing, and before the end event. 

Formally, a durative action \python{|\class{DurativeAction}|} has a string of input {\em parameters} \python{|\kwarg{params}|}, a list of {\em start conditions} \python{|\kwarg{start\_cond}|} that must hold prior to starting, a list of {\em start effects} \python{|\kwarg{start\_effects}|} that take effect after starting, a list of {\em ongoing conditions} \python{|\kwarg{over\_cond}|} that must remain true while the action is ongoing (exclusively), a list of {\em end conditions} \python{|\kwarg{end\_cond}|} that must hold prior to ending, a list of {\em end effects} \python{|\kwarg{end\_effects}|} that take effect after ending, and a {\em duration} \python{|\kwarg{duration}|} that specifies the minimum time elapsed between starting and ending the action.

In our running bimanual domain, the key durative action is \python{move}, which models arm motion where the arm may or may not be holding an object.
The \python{move} durative action has parameters for the arm \python{"?arm"}, start configuration \python{"?q1"}, trajectory \python{"?t"}, and end configuration \python{"?q2"}.
Its start conditions are that \python{"?t"} is a valid \python{Motion} (see Section~\ref{sec:streams}) that connects \python{"?q1"} and \python{"?q2"} and that \python{"?arm"} starts at configuration \python{"?q1"}. 
Critically, its start effect is that the \python{"?arm"} is no longer at \python{"?q1"} but instead is ``at'' trajectory \python{"?t"}, meaning that it is somewhere along the trajectory.
The \python{move} action does not have any end conditions, and its end effect is that the robot is at configuration \python{"?q2"}.
Finally, its duration is a numeric function of the trajectory \python{"?t"}.
Here, trajectories \python{"?t"} can be both long multi-second paths as well as short steering segments that are a part of a larger roadmap and executed via successive \python{move} actions.

\begin{footnotesize}
\begin{minted}{python}
move = |\class{DurativeAction}|(|\kwarg{params}|="?arm ?q1 ?t ?q2",
  |\kwarg{start\_cond}|=[Motion("?arm ?q1 ?t ?q2"), 
              At("?arm")=="?q1"],
  |\kwarg{start\_eff}|=[At("?arm")<="?t"],
  |\kwarg{over\_cond}|=collision_cond,
  |\kwarg{end\_cond}|=[], |\kwarg{end\_eff}|=[At("?arm")<="?q2"],
  |\kwarg{duration}|=Duration("?t"))
\end{minted}
\end{footnotesize}

In order to schedule with \python{move}, we need to evaluate the static function \python{Duration("?t")}.
Because we can't declare its output for infinitely many possible trajectories upfront,
we leverage an advantage of building off a programming language by defining \python{Duration} {\em procedurally} as a Python procedure rather than declaratively.
Namely, we extend \python{|\class{Function}|} to also take an optional Python function \python{|\kwarg{fn}|} that provides the unique, static value of the function.

\begin{footnotesize}
\begin{minted}{python}
Duration = |\class{Function}|("?t", |\kwarg{fn}|=...)
\end{minted}
\end{footnotesize}

The key conditions that ground \python{move} in robotics are its ongoing conditions \python{collision_cond}, which ensure that no collisions occur while \python{"?arm"} is moving along trajectory \python{"?t"}.
We use Python to decompose \python{collision_cond} into negated ($\sim$) {\em nested} arm-object collisions (\python{ObjCollision}) and arm-arm collisions (\python{ArmCollision}) that iterate over the movable objects \python{[obj1, obj2]} and robot arms \python{[arm1, arm2]}, consuming the current object or arm state.
Both of these procedural predicates take a Python Boolean function \python{|\kwarg{test}|} that returns \python{True} if, in the case of \python{ArmCollision}, the swept volume induced by trajectory \python{"?t1"} 
intersects with the swept volume induced by trajectory \python{"?t2"}.

\begin{footnotesize}
\begin{minted}{python}
ObjCollision = |\class{Predicate}|("?t ?p", |\kwarg{test}|=...)
ArmCollision = |\class{Predicate}|("?t1 ?t2", |\kwarg{test}|=...)
collision_cond = [|$\sim$|ObjCollision("?t",Pose(o)) 
  for o in [obj1,obj2]]+[|$\sim$|ArmCollision("?t",At(a)) 
  for a in [arm1,arm2]]
\end{minted}
\end{footnotesize}

\subsection{Samplers as Streams}\label{sec:streams}

In order to instantiate the \python{pick}, \python{place}, and \python{move} actions in Sections~\ref{sec:instantaneous}
and \ref{sec:durative}, we need to be able to generate possibly arbitrarily-many continuous parameter values for them.
Additionally, combinations of these parameter values must satisfy the static predicates \python{Kin}, which holds if forward kinematics from configuration \python{"?q"} times grasp pose \python{"?g"} results in placement pose \python{"?p"}, and motion \python{Motion}, which holds if trajectory \python{"?t"} avoids collisions with static objects, respects dynamical joint limits, and connects configurations \python{"?q1"} and \python{"?q2}.

\begin{footnotesize}
\begin{minted}{python}
Grasp = |\class{Predicate}|("?arm ?obj ?g", 
  |\kwarg{cond}|=[Arm("?arm"), Object("?obj")])
Kin = |\class{Predicate}|("?arm ?q ?obj ?g ?p", 
  |\kwarg{cond}|=[Conf("?arm ?q"), Placement("?obj ?p"),
        Grasp("?arm ?obj ?g")])
Motion = |\class{Predicate}|("?arm ?q1 ?t ?q2", 
  |\kwarg{cond}|=[Conf("?arm ?q1"), Conf("?arm ?q2")])
\end{minted}
\end{footnotesize}

We build on the PDDLStream framework~\cite{garrett2020PDDLStream} by extending a typical finite action language~\cite{mcdermott1998pddl} with {\em streams} \python{|\class{Stream}|}, procedural conditional generators that consume existing constants and generate new ones that satisfy specific predicates.
We can derive a stream from a predicate by partitioning its arguments into inputs \python{|\kwarg{inp}|} and outputs.
This also partitions the predicate's conditions \python{|\kwarg{cond}|} into input conditions, which the input values must satisfy to call the stream, and output conditions, which all generated values are guaranteed to satisfy.
For example, the \python{grasps} stream generates grasp poses \python{"?q"} for a given arm \python{"?arm"} and object \python{"?obj"}, the inverse kinematics stream \python{ik_qs} generates configurations \python{"?q"} that satisfy kinematics with input grasp and placement poses \python{"?g"}, \python{"?p"}, and the motion planner stream \python{motions} plans trajectories \python{"?t"} that connect configurations \python{"?q1"}, \python{"?q2"}.
In Section~\ref{sec:batch}, we discuss how these samplers can be GPU accelerated.

\begin{footnotesize}
\begin{minted}{python}
grasps = |\class{Stream}|(Grasp, |\kwarg{inps}|="?arm ?obj", |\kwarg{gen}|=...)
ik_qs = |\class{Stream}|(Kin, |\kwarg{inps}|="?arm ?obj ?g ?p", ...)
motions = |\class{Stream}|(Motion, |\kwarg{inps}|="?arm ?q1 ?q2", ...)
\end{minted}
\end{footnotesize}

\approach{} algorithms can call each stream's conditional generator \python{|\kwarg{gen}|} to enumerate new constants and, in turn, pass these new constants to other streams and functions.
A typical sequence of operations in our example is as follows, where the underlined values are generated by streams.

\begin{footnotesize}
\begin{minted}{python}
|\underline{g}|, = next(grasps.|\kwarg{gen}|(arm1))
|\underline{q}|, = next(ik_qs.|\kwarg{gen}|(arm1, obj1, |\underline{g}|, p1))
|\underline{t}|, = next(motions.|\kwarg{gen}|(arm1, q1, |\underline{q}|))
duration = Duration.|\kwarg{fn}|(|\underline{t}|)
collision_cond = |\kw{not}| ObjCollision.|\kwarg{test}|(|\underline{t}|,p1) 
  and |\kw{not}| ArmCollision.|\kwarg{test}|(|\underline{t}|,q2)
\end{minted}
\end{footnotesize}

\subsection{\approach{} Problems}\label{sec:problem}

Finally, we formalize \approach{} problems.
\begin{defn}\label{defn:problem}
    A \approach{} {\em problem} $\langle {\cal I}, {\cal G}, {\cal A}, {\cal S} \rangle$ is defined by an initial state ${\cal I}$ of function assignments, a list of conjunctive goal conditions ${\cal G}$, a set of durative (and instantaneous) actions ${\cal A}$, and a set of streams ${\cal S}$.
\end{defn}

\begin{defn}\label{defn:schedule}
A solution to a \approach{} problem 
is a {\em schedule} 
$\tau = [\langle t_1, a_1(x_1), t_1' \rangle, ..., \langle t_n, a_n(x_n), t_n' \rangle]$ that respects action and goal conditions as well as has start and end times $t_i, t_i'$ and action instances $a_i(x_i)$, where $t_i, t_i' \in [0, \infty)$, $a_i \in {\cal A}$, $t_i + a_i(x_i).\var{duration} \leq t_i'$, for $i \in [n]$.
\end{defn}
\noindent The objective of \approach{} planning is to find a schedule $\tau$ with minimal {\em makespan} $t_* {\equiv} \max_{i \in [n]} t_i'$, i.e. duration. %

\section{SCHEDULESTREAM ALGORITHMS}\label{sec:method}

We now present algorithms that solve \approach{} problems without any domain-specific additional inputs. 
Our algorithms alternate between {\em scheduling} and stream phases.

\subsection{Eager Stream Scheduling} \label{sec:eager}

The simplest \approach{} algorithm is \proc{eager-stream} (Algorithm~\ref{alg:eager}), which iteratively attempts to solve a finite scheduling problem using subroutine \proc{schedule} (Section~\ref{sec:schedule}) with the current set of constants in an {\em augmented} initial state $I$.
Upon failing to find a schedule $\tau$, \proc{eager-stream} 
instantiates streams $s \in {\cal S}$ into stream instances $s(x)$, where $x$ are constants in $I$.
For each stream instance $s(x)$, the conditional generator of its stream $s.\var{gen}$ is called using constants $x$ as inputs, producing new output constants $y$ or $\kw{None}$ if none exist.
If $y \neq \kw{None}$, the stream output $s(x)(y)$ is added to {\em stream plan} $\psi$, which records a sequence of stream operations that support the ultimate solution schedule $\tau$.
Finally, the augmented initial state $I$ is updated to include function assignments $p(x, y) \gets \kw{True}$ for the stream output predicates $s.\var{out\_cond}$, which induces a new finite scheduling problem on the next iteration.

\begin{algorithm}[h]
    \caption{Eager Stream Scheduling}
    \label{alg:eager}
    \begin{algorithmic}[1]
    \begin{footnotesize}
        \Procedure{eager-stream}{${\cal I}, {\cal G}, {\cal A}, {\cal S}$}
            \State $I \gets \kw{copy}({\cal I})$; $\psi \gets [\;]$ \Comment{Augmented initial state, stream plan}
            \While{\True} %
                \State $\tau \gets \proc{schedule}(I, {\cal G}, {\cal A})$ \Comment{Solve a scheduling subproblem}
                \If{$\tau \neq \None$}
                    \State \Return $\psi, \tau$ \Comment{Return stream plan and schedule}
                \EndIf
                \For{$s(x) \in \proc{instantiate}(I, {\cal S})$}
                    \State $y \gets \kw{next}(s.\var{gen}(x))$ \Comment{Generate a new stream output}
                    \If{$y \neq \None$ \kw{and} $s(x)(y) \notin \psi$}
                        \State $\psi \gets \psi + [s(x)(y)]$ \Comment{Record new stream output}
                        \For{$p \in s.\var{out\_cond}$}
                            \State $I[p(x,y)] \gets \kw{True}$ \Comment{Augment the initial state} %
                        \EndFor
                    \EndIf
                \EndFor
            \EndWhile
        \EndProcedure
    \end{footnotesize}
    \end{algorithmic}
\end{algorithm}

\subsection{Schedule Subroutine}\label{sec:schedule}

A key subroutine in \proc{eager-stream} is $\proc{schedule}$, which solves a finite planning \& scheduling problem defined by initial state $I$, goal conditions ${\cal G}$, and durative actions ${\cal A}$.
This requires not only selecting the action instances $a(x)$ required but also timing them into a schedule $\tau$, exceeding the capabilities of classical planning algorithms~\cite{hoffmann2001ff}.
Building on ideas in temporal planning~\cite{eyerich09icaps}, we reduce scheduling to sequential planning of the {\em start} and {\em end} of each applied action instance.
To do this, we compile each durative action $a$ into two instantaneous actions, a start action $a.\var{start}$ and an end action $a.\var{end}$, and solve a sequential planning problem for a plan $\pi$ for the order to start and end each action.

\begin{algorithm}[h]
    \caption{Finite Planning \& Scheduling}
    \label{alg:temporal}
    \begin{algorithmic}[1]
    \begin{footnotesize}
        \Procedure{schedule}{$I, {\cal G}, {\cal A}$}
        \State $A \gets \{\widehat{a}(x) \mid a(x) \in \proc{instantiate}(I, {\cal A}).\; \widehat{a} \in [a.\var{start}, a.\var{end}]\}$
        \State $\pi \gets \proc{search}(I, {\cal G}, A)$ \Comment{Sequential search}
        \If{$\pi = \kw{None}$} \Return $\kw{None}$ \EndIf
        \State $\pi \gets \proc{search}(I, {\cal G}, \pi)$ \Comment{Reschedule initial plan $\pi$}
        \State $\tau \gets [\;], t_* \gets 0$; $T \gets \{\;\}$\Comment{Schedule, current time, and starts}
        \For{$\widehat{a}(x) \in \pi$}
            \If{$\widehat{a} = a.\var{start}$} \Comment{Start action $a.\var{start}$}
                \State $T[a(x)] \gets t_*$ \Comment{Set start time}
            \Else \Comment{End action ($\widehat{a} = a.\var{end}$)}
                \State $\Delta t \gets a(x).\var{duration} - (t_* - T[a(x)])$ \Comment{Elapsed time}
                \State $t_* \gets t_* + \kw{max}(0, \Delta t)$ \Comment{New current time}
                \State $\tau \gets \tau + [\langle T[a(x)], a(x), t_* \rangle]$ \Comment{Add action to schedule}
            \EndIf
        \EndFor
        \State \Return $\tau$ \Comment{Return the schedule $\tau$ derived from plan $\pi$}
        \EndProcedure
    \end{footnotesize}
    \end{algorithmic}
\end{algorithm}

Let action \python{a} be a durative action.
First, we create a predicate \python{Ongoing} that is true if action \python{a} is currently being executed and a function \python{Ongoing} that tracks the minimum time remaining in order to terminate the action.
Both have arguments corresponding to the original action parameters.
We define the {\em start action} \python{start_a} to inherit the start conditions and effects of action \python{a}, add a condition that \python{a} is not currently ongoing, and add effects that \python{a} is now ongoing and its remaining time is equal to the duration of the action.

\begin{footnotesize}
\begin{minted}{python}
Ongoing = |\class{Predicate}|(|\kwarg{args}|=a.|\kwarg{params}|)
Remaining = |\class{Function}|(|\kwarg{args}|=a.|\kwarg{params}|)
start_a = |\class{Action}|(|\kwarg{params}|=a.|\kwarg{params}|, 
  |\kwarg{cond}|=a.|\kwarg{start\_cond}| + [|$\sim$|Ongoing(a.|\kwarg{params}|),
    OverCondition("#s")],
  |\kwarg{eff}|=a.|\kwarg{start\_eff}| + [Ongoing(a.|\kwarg{params}|),
    Remaining(a.|\kwarg{params}|)<=a.|\kwarg{duration}|])
\end{minted}
\end{footnotesize}

\noindent
The {\em end action} \python{end_a} similarly inherits the end conditions and effects of action \python{a}, adds a condition that \python{a} is currently ongoing, and an effect that afterward \python{a} is no longer ongoing.
The {\em cost} of \python{end_a} is time remaining, which can be seen as advancing time forward by this amount.
Additionally, we leverage Python by defining a procedural condition \python{OverCondition} that checks whether the ongoing conditions \python{a2.|\kwarg{over\_cond}|} for each ongoing action \python{a2} are still satisfied 
as well as a procedural effect \python{ElapseEffect} that updates the \python{Remaining} duration for each ongoing action \python{a2} by subtracting the remaining duration for \python{a}.

\begin{footnotesize}
\begin{minted}{python}
OverCondition = |\class{Condition}|(|\kwarg{args}|="#s", |\kwarg{fn}|=...)
ElapseEffect = |\class{Effect}|(|\kwarg{args}|="#s ?d", |\kwarg{fn}|=...)
end_a = |\class{Action}|(|\kwarg{params}|=a.|\kwarg{params}|, 
  |\kwarg{cond}|=a.|\kwarg{end\_cond}| + [Ongoing(a.|\kwarg{params}|),
    OverCondition("#s")],
  |\kwarg{eff}|=a.|\kwarg{end\_eff}| + [|$\sim$|Ongoing(a.|\kwarg{params}|), 
    ElapseEffect("#s", Remaining(a.|\kwarg{params}|)),
    Remaining(a.|\kwarg{params}|)<=0.],
  |\kwarg{cost}|=Remaining(a.|\kwarg{params}|))
\end{minted}
\end{footnotesize}

Now, we implement \proc{schedule} (Algorithm~\ref{alg:temporal}) by first solving a \proc{search} problem with start and end actions to produce a sequential plan $\pi$ and then extracting a schedule $\tau$ that corresponds to that plan.
First, it instantiates the start and end actions for each durative action.
Then, it calls a generic search algorithm \proc{search} to find a sequential plan $\pi$, minimizing total end action cost (i.e. time).
We implement \proc{search} with a lazy weighted A$^*$ heuristic search using a modified version of the Fast-Forward (FF) heuristic~\cite{hoffmann2001ff} that ignores procedural and nested conditions.
If it produces a plan, we call \proc{search} again using just the action instances in $\pi$, with weight $w = 1$, in order to locally optimize the plan.
For start instance $a.\var{start}(x)$ in the plan, it records the start time as the current time $t_*$.
For end instances $a.\var{start}$, it computes the remaining time $\Delta t$ that must elapse to terminate the action.
It adds this to the current time $t_*$ or zero if $\Delta t < 0$.
Finally, \proc{schedule} returns the schedule $\tau$.

\subsection{Lazy Stream Scheduling}\label{sec:lazy}

\begin{figure*}[h]
    \centering
    \includegraphics[width=\textwidth]{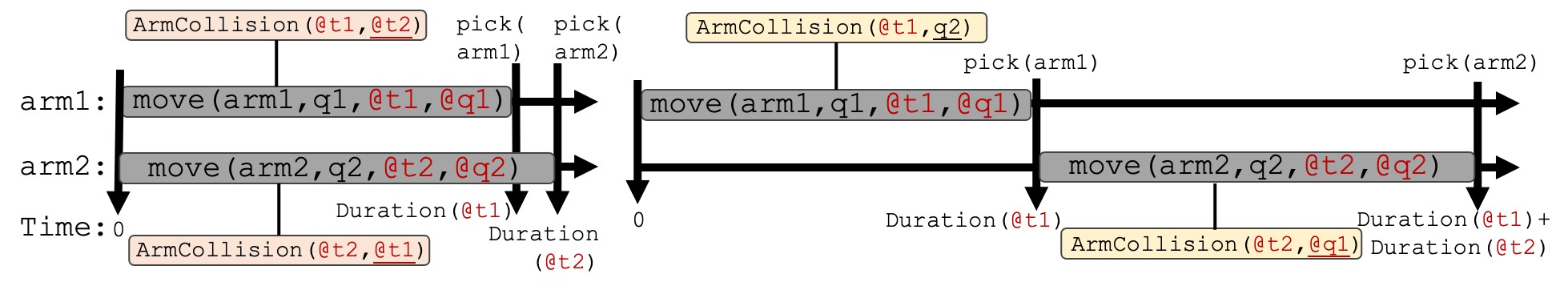}
    \caption{\small{\textbf{Example Schedules.} 
    {\em Left:} a solution schedule for {\it Example 1} in Figure~\ref{fig:bimanual}. Each arm is able to execute a move action in parallel because the trajectories \python{"@t1"}, \python{"@t2"} do not collide.
    {\em Right:} a solution schedule for {\it Example 2}. Although trajectories \python{"@t1"}, \python{"@t2"} collide, 
    the trajectory-configuration pairs \python{"@t1"}, \python{q2} and \python{"@t2"}, \python{"@q1"} do not, admitting a serial solution that has a larger makespan. %
    }}
    \label{fig:example}
\vspace{-20px}
\end{figure*}

Although the \proc{eager-stream} algorithm in Section~\ref{sec:eager} will provably solve \approach{} problems if a solution exists,
it will often unnecessarily call many streams not required to solve the problem, 
and streams for robotics domains involve, for example, expensive inverse kinematics operations.
Thus, building on prior work~\cite{garrett2020PDDLStream}, we propose a {\em lazy} stream scheduling algorithm that saves time by first identifying which streams could be useful for calling them.
The key insight is to temporarily replace a stream's conditional generator \python{s.|\kwarg{gen}|} with a version that creates unique placeholder constants (here denoted as strings with the prefix \python{"@"}) and plan with these inexpensive values first.
Later, an algorithm will retrace which placeholder constants are used and call the streams associated with those constants.
In the running example in Figure~\ref{fig:bimanual}, the lazy stream calls will form the following stream plan $\psi$, which here is a serialized computation graph of stream instances and outputs.

\begin{footnotesize}
\begin{minted}{python}
stream_plan = [grasps(arm1)("@g1"), ...
 ik_qs(arm1,obj1,"@g1",p1)("@q1"), ...
 motions(arm1,q1,"@q1")("@t1"), ...]
\end{minted}
\end{footnotesize}

\noindent From these lazy constants, we can instantiate lazy durative action instances that will be instantiated in \proc{schedule}.

\begin{footnotesize}
\begin{minted}{python}
actions = {..., 
 move(arm1,q1,"@t1","@q1"),pick(arm1,"@q1","@g1",p1), 
 move(arm2,q2,"@t2","@q2"),pick(arm2,"@q2","@g2",p2)}
\end{minted}
\end{footnotesize}

\noindent When evaluating functions involving lazy constants, we use the function's default value, which is \python{False} for collision predicates and \python{0.} for numeric duration functions.

\begin{algorithm}[h]
    \caption{Lazy Stream Scheduling}
    \label{alg:lazy}
    \begin{algorithmic}[1]
    \begin{footnotesize}
        \Procedure{lazy-stream}{${\cal I}, {\cal G}, {\cal A}, {\cal S}$}
        \State $P \gets \emptyset$ \Comment{Initialize stream plan, schedule skeleton pairs}
        \While{\True}
            \For{$s \in {\cal S}$}
                \State $s.\var{gen} \gets \proc{lazy-generator}(s, P)$ \Comment{Lazy output creation}
            \EndFor
            \State $\psi', \tau \gets \proc{eager-stream}({\cal I}, {\cal G}, {\cal A}, {\cal S})$ \Comment{Reduce to eager}
            \If{$\tau \neq \kw{None}$}
                \State $\psi \gets \proc{retrace-streams}({\cal I}, \psi', \proc{preimage}(\tau, {\cal G}))$
                \State $P \gets P \cup \{\langle \psi, \tau \rangle\}$ \Comment{Add schedule skeleton}
            \EndIf
            \For{$\langle \psi, \tau \rangle \in P$}
            \State $Y \gets \{\;\}$ \Comment{Initialize stream output mapping}
            \For{$s(x)(y) \in \psi$}
                \State $x' \gets \proc{substitute}(x, Y)$ \Comment{Remap input values}
                \State $y' \gets \kw{next}(s.\var{gen}(x'))$ \Comment{Generate a stream output}
                \If{$y' = \None$}
                    $Y \gets \kw{None}$; \kw{break} \Comment{Failure}
                \EndIf
                \State $Y[y] \gets y'$ \Comment{Update the stream output mapping}
            \EndFor
            \If{$Y \neq \None$}
                \State \Return $\proc{substitute}(\psi, Y), \proc{substitute}(\tau, Y)$
            \EndIf
            \EndFor
        \EndWhile
        \EndProcedure
    \end{footnotesize}
    \end{algorithmic}
\end{algorithm}

Algorithm~\ref{alg:lazy} gives the pseudocode for the lazy stream scheduling algorithm \proc{lazy-stream}.
It maintains a set of identified {\em schedule skeletons} $P$, schedules $\tau$ with lazy constants along with stream plans $\psi$ that could bind values for those constants.
On each iteration, \proc{lazy-stream} replaces each stream conditional generator $s.\var{gen}$ with a version that's lazy for stream instances $s(x)$ not on a stream plan $\psi$ in $P$.
It then calls \proc{eager-stream} (Algorithm~\ref{alg:eager}) to identify a new stream plan $\psi'$ and schedule skeleton $\tau$.
From $\psi'$, it retraces a minimal subsequence $\psi$ that's sufficient for generating values for the lazy constants on $\tau$.
At the end of each iteration, \proc{lazy-stream} evaluates each stream instance $s(x)$ on an identified stream plan $\psi$, attempting to replace each lazy output $y$ with a real one $y'$ in a map of bindings $Y$.
If this process is successful, it replaces all lazy constants in $\psi$ and $\tau$ with their new values in $Y$ and returns this pair.

\subsection{Illustrated Example}\label{sec:example}

Revisiting our running example in Figure~\ref{fig:bimanual}, we now discuss the solutions to each of the three problem instances for the task for \python{arm1} to hold object \python{obj1} and \python{arm2} to hold object \python{obj2}.
We will consider schedules computed by \proc{lazy-stream} (Section~\ref{sec:lazy}), where configurations \python{"@q1"} and  \python{"@q2"} are lazy outputs of stream \python{ik_qs} and trajectories \python{"@t1"} and \python{"@t2"} are lazy outputs of stream \python{motions}.  
Figure~\ref{fig:example} ({\em left}) displays a schedule for {\em Problem 1}, where the arms are able to fully parallelize motion because trajectories \python{"@t1"} and \python{"@t2"} are never in collision.
Figure~\ref{fig:example} ({\em right}) displays a schedule for {\em Problem 2}, where the arms must move serially because trajectories \python{"@t1"} and \python{"@t2"} are in collision; however, \python{"@t1"} does not collide with \python{q2} and \python{"@t2"} does not collide with \python{"@q1"}.
This schedule uses the same durative action instances as in {\em Problem 1} but in a different start and end order.
Finally, neither of these schedules solve {\em Problem 3} because \python{pick} configurations \python{"@q1"} and \python{"@q2"} collide.
As a result, the planner must introduce an additional \python{move} action to move \python{arm1} out of the way to make room for the other \python{arm2}. %

\section{GPU-ACCELERATED STREAMS} %
\label{sec:batch}

Another challenge of performing temporal instead of sequential TAMP is the need to check collisions between pairs of moving arms instead of a single moving arm against a stationary one.
Moreover, the \python{ArmCollision} procedural predicate in the \python{move} action's conditions (Section~\ref{sec:durative}) checks for collisions between all pairs of configurations along two trajectories in the event that another arm \python{arm} is moving, namely when \python{At(arm)} is a trajectory.
As an example, consider the usage of \python{ArmCollision} in the schedule for {\em Example 1} versus {\em Example 2} in Figure~\ref{fig:example}.
In {\em Example 1}, it computes collisions between lazy trajectories \python{"@t1"}, \python{"@t2"} whereas in {\em Example 2}, it computes collisions between trajectory and configuration pairs.
Sequential schedules avoid a quadratic number of collision checks; however, they have higher makespans than parallel schedules.

To ease this computational burden, we use GPU acceleration to batch together expensive collision checking, forward kinematics, inverse kinematics, and motion planning stream calls in Section~\ref{sec:streams}.
Our use of GPU acceleration for multi-arm planning goes beyond that of cuTAMP~\cite{shen2025cutamp}, which only considers single-arm TAMP.
Specifically, we build on the cuRobo~\cite{curobo_report23} custom CUDA kernels.
We represent each robot as a union of inflated inscribed spheres derived from its meshes, which we greedily sample for a given computation budget~\cite{inui2016shrinking}.
For moving object collisions, we perform asymmetric sphere-mesh collision checks where we transform the sampled swept volume of spheres along a arm trajectory into the frame of an object at a batch of poses.
For moving arm collisions, we cast pairwise arm collisions as a batch sphere-sphere self-collision check for a two arm composite robot.

\begin{figure*}[ht!]
    \centering
    \includegraphics[trim={0cm 9cm 2cm 12cm},clip,width=0.435\linewidth]{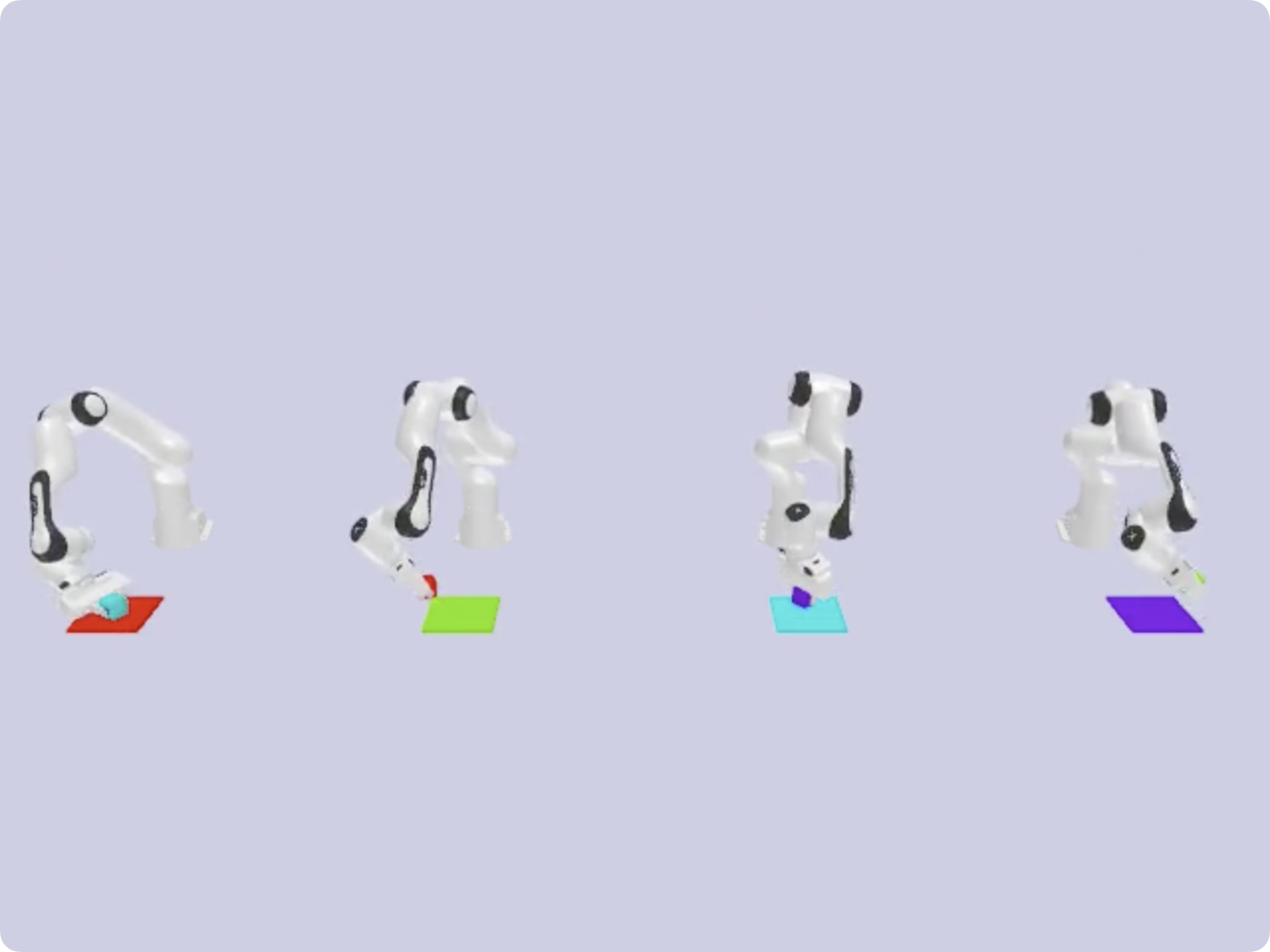}
    \includegraphics[trim={12cm 10cm 11cm 4cm},clip,width=0.145\linewidth]{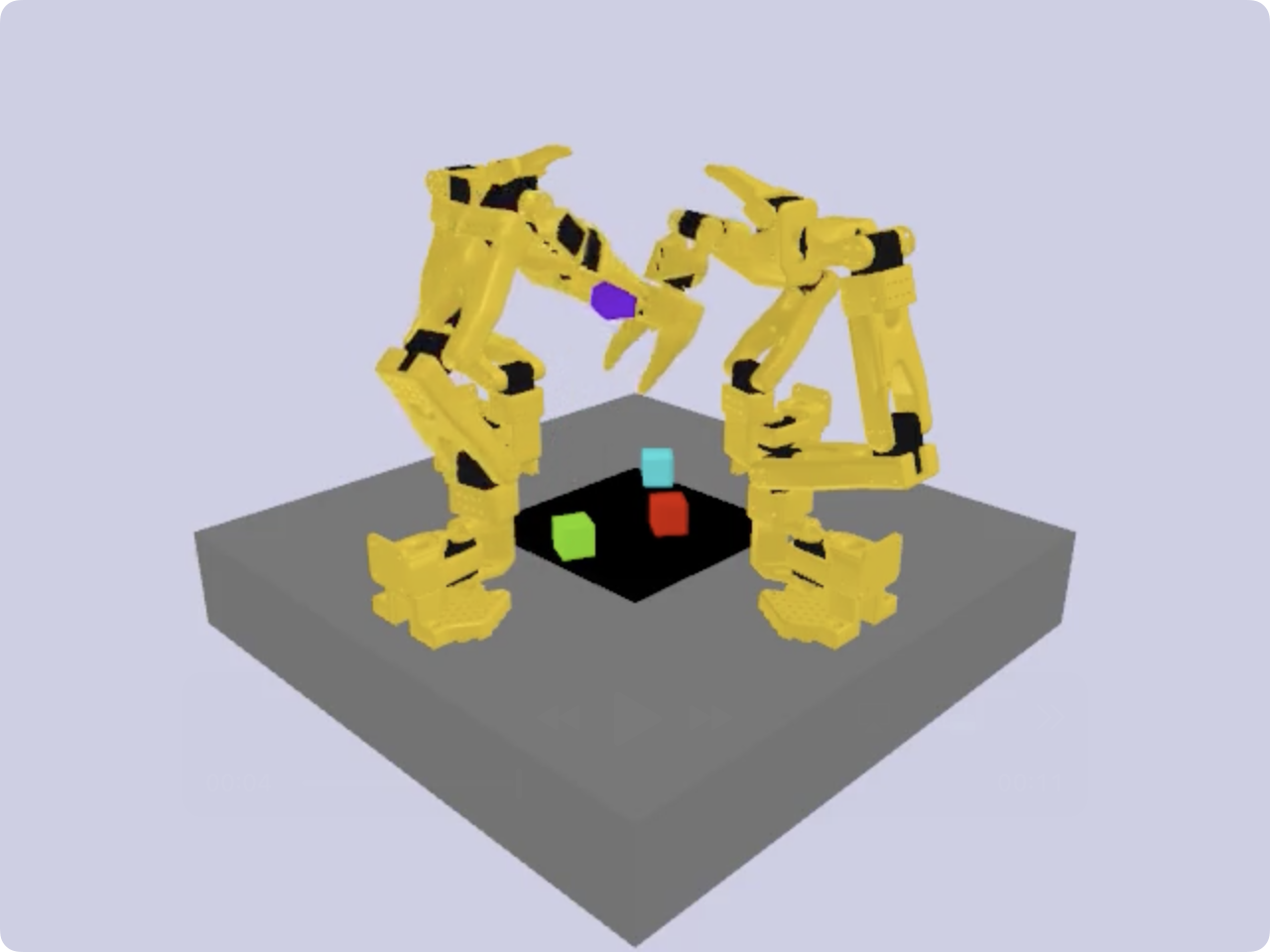}
    \includegraphics[trim={9cm 10cm 11cm 10cm},clip,width=0.23\linewidth]{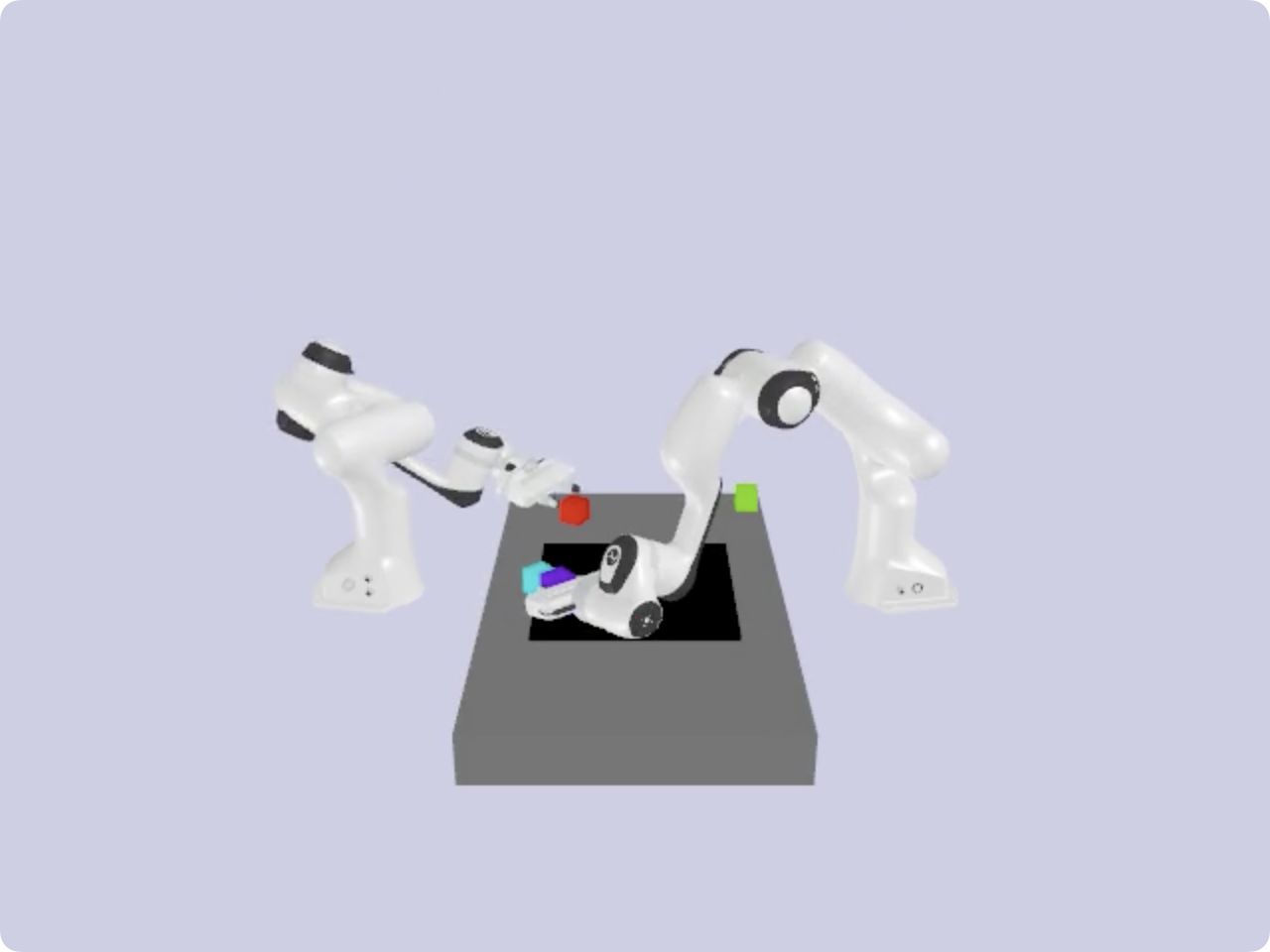}
    \includegraphics[trim={10cm 11cm 5cm 0cm},clip,width=0.17\linewidth]{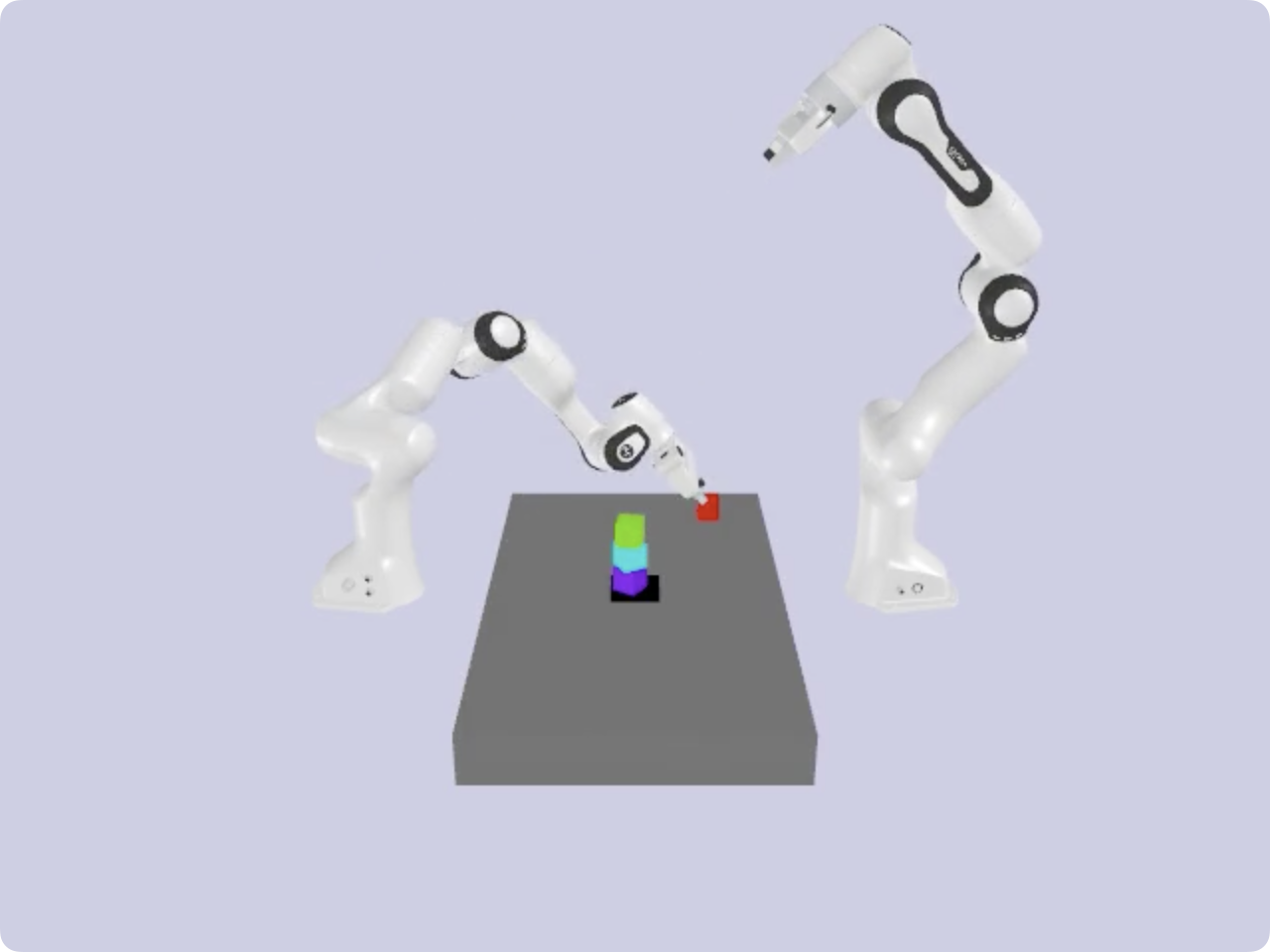}
    \caption{\small{\textbf{Simulated Experiments.} %
    The ``Franka Hold Any 4'', ``SO100 Hold Any 4'', ``Franka Pack 4'', and ``Franka Stack 4'' tasks.}}
    \label{fig:tasks}
    \vspace{-2em}
\end{figure*}

\section{RESULTS}\label{sec:result}

We analyzed the performance of \approach{} compared to ablations representative of prior works in simulation (Section~\ref{sec:simulated}) and tested it in the real-world (Section~\ref{sec:real}).

\begin{figure}[h]
\centering
\setlength{\tabcolsep}{3pt}
\begin{footnotesize}
\begin{tabular}{|l|rr|rr|rr|rr|}
\toprule
\textbf{Algorithm:} & \multicolumn{2}{c|}{\textit{Sequential}} & \multicolumn{2}{c|}{\textit{Hierarchical}} & \multicolumn{2}{c|}{\textit{Ours}} & \multicolumn{2}{c|}{\textit{Ours{+}GPU}} \\
\midrule
\textbf{Task} &
\% & Time &
\% & Time &
\% & Time &
\% & Time \\
\midrule
Franka Assigned 1 & 100 & 0.4 & 100 & 0.4 & 100 & 0.4 & 100 & 0.1 \\
Franka Assigned 2 & 100 & 1.7 & 100 & 1.7 & 100 & 1.7 & 100 & 0.4 \\
Franka Assigned 3 & 100 & 3.2 & 100 & 4.9 & 100 & 5.0 & 100 & 0.7 \\
Franka Assigned 4 & 100 & 8.2 & 100 & 14.0 & 100 & 14.0 & 100 & 1.4 \\
\midrule
Franka Any 2 & 100 & 2.4 & 42 & 35.6 & 100 & 2.4 & 100 & 0.9 \\
Franka Any 3 & 100 & 5.7 & 15 & 51.9 & 100 & 7.8 & 100 & 2.5 \\
Franka Any 4 & 100 & 22.1 & 5 & 57.9 & 99 & 31.6 & 100 & 6.8 \\
\midrule
SO100 Any 2 & 95 & 2.4 & 52 & 20.3 & 95 & 23.7 & 95 & 1.0 \\
SO100 Any 3 & 99 & 5.9 & 24 & 58.1 & 58 & 115.4 & 100 & 2.3 \\
SO100 Any 4 & 100 & 9.1 & 9 & 101.4 & 18 & 244.2 & 99 & 4.4 \\
\midrule
Franka Pack 2 & 100 & 4.0 & 99 & 6.3 & 100 & 4.6 & 100 & 1.0 \\
Franka Pack 3 & 100 & 6.7 & 76 & 21.5 & 100 & 8.9 & 100 & 1.5 \\
Franka Pack 4 & 90 & 12.4 & 41 & 40.2 & 96 & 18.2 & 99 & 2.3 \\
\midrule
Franka Stack 2 & 100 & 3.8 & 99 & 5.7 & 100 & 4.0 & 100 & 0.9 \\
Franka Stack 3 & 100 & 6.4 & 89 & 12.3 & 99 & 7.5 & 100 & 1.4 \\
Franka Stack 4 & 100 & 9.1 & 80 & 19.9 & 98 & 12.3 & 100 & 2.4 \\
\midrule
\textbf{Average} & \textbf{99} & 6.5 & 64 & 28.3 & 91 & 31.4 & \textbf{99} & \textbf{1.9} \\
\bottomrule
\end{tabular}
\end{footnotesize}
\captionof{table}{
\textbf{Success Rate and First Solution Time}. \textit{Hierarchical} fails to solve several tasks because it does not backtrack,
\textit{Ours} has a lower success rate and higher runtime than its GPU-accelerated counterpart \textit{Ours + GPU}.
}
\label{tab:success}
\vspace{-20px}
\end{figure}

\begin{figure}[h] %
\centering
\setlength{\tabcolsep}{3pt}
\begin{footnotesize}
\begin{tabular}{|l|rr|rr|rr|rr|}
\toprule
\textbf{Algorithm:} & \multicolumn{2}{c|}{\textit{Sequential}} & \multicolumn{2}{c|}{\textit{Hierarchical}} & \multicolumn{2}{c|}{\textit{Ours}} & \multicolumn{2}{c|}{\textit{Ours{+}GPU}} \\
\midrule
\textbf{Task} &
$t_*^1$ & $t_*^\infty$ &
$t_*^1$ & $t_*^\infty$ &
$t_*^1$ & $t_*^\infty$ &
$t_*^1$ & $t_*^\infty$ \\
\midrule
Franka Assigned 1 & 0.7 & 0.6 & 0.7 & 0.6 & 0.7 & 0.6 & 0.7 & 0.6 \\
Franka Assigned 2 & 1.4 & 1.2 & 0.8 & 0.6 & 0.8 & 0.6 & 0.8 & 0.6 \\
Franka Assigned 3 & 2.1 & 1.8 & 0.8 & 0.6 & 0.8 & 0.6 & 0.8 & 0.6 \\
Franka Assigned 4 & 2.8 & 2.4 & 0.8 & 0.7 & 0.8 & 0.8 & 0.8 & 0.6 \\
\midrule
Franka Any 2 & 1.4 & 1.2 & 0.7 & 0.6 & 0.7 & 0.6 & 0.8 & 0.6 \\
Franka Any 3 & 2.1 & 1.8 & 0.8 & 0.7 & 0.8 & 0.7 & 0.8 & 0.6 \\
Franka Any 4 & 2.9 & 2.6 & 0.9 & 0.7 & 0.8 & 0.8 & 0.8 & 0.6 \\
\midrule
SO100 Any 2 & 3.9 & 3.8 & 2.0 & 2.0 & 2.3 & 2.2 & 2.1 & 2.0 \\
SO100 Any 3 & 5.8 & 5.7 & 2.0 & 2.0 & 2.8 & 2.8 & 2.8 & 2.1 \\
SO100 Any 4 & 7.8 & 7.6 & 2.0 & 2.0 & 2.3 & 2.3 & 3.6 & 2.4 \\
\midrule
Franka Pack 2 & 2.6 & 2.1 & 1.4 & 1.2 & 1.7 & 1.2 & 1.5 & 1.1 \\
Franka Pack 3 & 4.0 & 3.3 & 2.7 & 2.3 & 3.0 & 2.3 & 3.0 & 2.2 \\
Franka Pack 4 & 5.4 & 4.6 & 2.9 & 2.5 & 4.3 & 3.4 & 3.9 & 2.8 \\
\midrule
Franka Stack 2 & 2.5 & 2.2 & 1.8 & 1.6 & 1.9 & 1.6 & 1.9 & 1.5 \\
Franka Stack 3 & 4.0 & 3.5 & 3.1 & 2.8 & 3.4 & 2.8 & 3.2 & 2.6 \\
Franka Stack 4 & 5.3 & 4.9 & 4.4 & 4.0 & 4.6 & 4.2 & 4.5 & 3.8 \\
\midrule
\textbf{Average} & 3.4 & 3.1 & 1.7* & 1.5* & 1.9* & 1.6* & \textbf{2.0} & \textbf{1.5} \\
\bottomrule
\end{tabular}
\end{footnotesize}
\captionof{table}{
\textbf{First and Final Makespans}. By making use of multiple arms and parallel execution, \textit{Ours} (\textit{+ GPU}) produces schedules on average half as long as \textit{Sequential}.
}
\label{tab:makespan}
\vspace{-20px}
\end{figure}

\subsection{Simulated Experiments}\label{sec:simulated}

We are interested in evaluating the success rate, runtime, and solution quality of \approach{} algorithms compared to prior ones.
We compared four algorithms that consume the same \approach{} problem formulation.
\begin{enumerate}
    \item \textit{Sequential}: 
    an ablation that reduces durative actions to traditional actions and thus performs sequential planning. This is representative of traditional TAMP approaches, namely PDDLStream~\cite{garrett2020PDDLStream} algorithms.
    \item \textit{Hierarchical}: an ablation that performs schedules once and attempts to satisfy the schedule in a strict hierarchy without backtracking. This is representative of multi-arm task {\em then} motion planning approaches~\cite{chen2022cooperative}.
    \item \textit{Ours}: \proc{lazy-stream} in Algorithm~\ref{alg:lazy}.
    \item \textit{Ours + GPU}: \proc{lazy-stream} with batched instead of sequential GPU-accelerated samplers in Section~\ref{sec:batch}.
\end{enumerate}

We experimented on five tasks (Figure~\ref{fig:tasks}), each with parameter $N$ that specifies the number of objects or robots, that test how the algorithms scale as $N$ increases.
\begin{enumerate}
    \item \textit{Franka Hold Assigned $N$}: $N$ Franka robots must hold an assigned one of $N$ objects on separate platforms.
    \item \textit{Franka Hold Any $N$}: $N$ Franka robots must each hold any one of $N$ objects on separate platforms. This is more challenging than ``Franka Hold $N$'' as algorithms must match robots to objects based on reachability.
    \item \textit{SO100 Hold Any $N$}: $N$ SO100 robots must each hold any one of $N$ objects on the same cluttered platform.
    \item \textit{Franka Pack $N$}: 2 Franka robots pack $N$ objects randomly on the floor into a tight region.
    \item \textit{Franka Stack $N$}: 2 Franka robots stack $N$ objects randomly on the floor on top of each other.
\end{enumerate}

We randomly sampled 100 problems per task, ran each algorithm in an anytime mode for 60 seconds, and recorded the discovery time and makespan of each identified solution.
Table~\ref{tab:success} lists the success rate (\%) and runtime to produce the {\em first} solution (Time) per task.
{\em Hierarchical} results in low success rates because sampled problems are not always downward refinable~\cite{bacchus1991downward}, particularly when only certain arms can reach certain objects.
As a result, performing scheduling without any grounding in physical robotic constraints is generally insufficient.
GPU acceleration clearly improves the success rate and decreases the runtime. %
Table~\ref{tab:makespan} lists the makespan of the first solution ($t_*^1$) and the best solution ($t_*^\infty$) after 60 seconds.
Because {\em Hierarchical} fails to solve many of the longer-horizon problems, their makespans are averaged across only the easier problems, resulting in a slightly lower average makespan (denoted by *).
As expected, {\em Sequential} produces much longer solutions than {\em Ours} because it does not exploit opportunities for parallelism.

\subsection{Real World Demonstration}\label{sec:real}

We validated \approach{} through 21 real-world demonstrations on the custom bimanual robot in Figure~\ref{fig:teaser}, which has two Kinova Gen3 arms.
We deployed \approach{} using a strategy similar to that of Curtis {\it et al.}~\cite{m0m}, where we used GroundingDINO~\cite{liu2023grounding} for open-world object detection and estimated object affordances from the segmented point cloud corresponding to each detection.
\approach{} was indeed able to plan simultaneous motion that makes use of both arms to efficiently complete 10 real-world tasks involving retrieving, sorting, packing, stacking, and regrasping.
See \href{https://schedulestream.github.io/}{website} for videos of the demonstrations as well as an open-source implementation of ScheduleStream.

\section{CONCLUSION}\label{sec:conclusion}

We introduced \approach{}, a new framework for planning \& scheduling with samplers that can be used to efficiently solve Task and Motion Planning \& Scheduling problems.
We showed that \approach{} algorithms result in higher success rates and lower makespans than sequential-only algorithms and demonstrated \approach{} on several real-world bimanual manipulation tasks.

\bibliographystyle{IEEEtran}
\bibliography{references}

\end{document}